\documentclass[preprint,11pt,3p]{elsarticle}

\usepackage{amssymb}
\usepackage{makeidx}  
\usepackage{graphicx}
\usepackage{array,colortbl,multirow,multicol,booktabs}
\usepackage[flushleft]{threeparttable}
\usepackage{natbib}
\usepackage{amsmath}
\usepackage{algorithm}
\usepackage{rotating}
\usepackage[noend]{algpseudocode}
\usepackage{paralist}
\usepackage{tabularx}

\usepackage{amsthm}

\usepackage{lineno}

\begin{document}

\begin{frontmatter}

\title{PRESISTANT: Learning based assistant for data pre-processing}

\author[label1,label2]{Besim Bilalli\corref{cor1}}
\ead{bbilalli@essi.upc.edu}
\cortext[cor1]{Corresponding author}

\author[label1]{Alberto Abell\'{o}}
\ead{aabello@essi.upc.edu}
\address[label1]{Universitat Polit\'{e}cnica de Catalunya, BarcelonaTech, Barcelona, Spain}

\author[label1]{Tom\`{a}s Aluja-Banet}
\ead{tomas.aluja@upc.edu}

\author[label2]{Robert Wrembel}
\ead{robert.wrembel@cs.put.poznan.pl}
\address[label2]{Poznan University of Technology, Poznan, Poland}

\begin{abstract}
Data pre-processing is one of the most time consuming and relevant steps in a data analysis process (e.g., classification task). A given data pre-processing operator (e.g., transformation) can have positive, negative or zero impact on the final result of the analysis. 
Expert users have the required knowledge to find the right pre-processing operators. However, when it comes to non-experts, they are overwhelmed by the amount of pre-processing operators and it is challenging for them to find operators that would positively impact their analysis (e.g., increase the predictive accuracy of a classifier). 
Existing solutions either assume that users have expert knowledge, or they recommend pre-processing operators that are only ``syntactically" applicable to a dataset, without taking into account their impact on the final analysis.
In this work, we aim at providing assistance to non-expert users by recommending data pre-processing operators that are ranked according to their impact on the final analysis. 
We developed a tool PRESISTANT, that uses Random Forests to learn the impact of pre-processing operators on the performance (e.g., predictive accuracy) of 5 different classification algorithms, such as J48, Naive Bayes, PART, Logistic Regression, and Nearest Neighbor. Extensive evaluations on the recommendations provided by our tool, show that PRESISTANT can effectively help non-experts in order to achieve improved results in their analytical tasks.




\end{abstract}


\end{frontmatter}



\section{Introduction}
Although machine learning algorithms have been around since the 1950s, their initial impact has been insignificant. With the increase of data availability and computing power, machine learning tools and algorithms are making breakthroughs in very diverse areas. Their success has raised the need for mainstreaming the use of machine learning, that is, engaging even non-expert users to perform data analysis. However, the multiple steps involved in the data analysis process render this process challenging. 

Data analysis as defined in~\cite{FayyadKDD96}, consists of data selection, data pre-processing, data mining, and evaluation or interpretation. A very important and time consuming step that marks itself out of the rest, is the data pre-processing step. Data pre-processing is challenging but at the same time has a heavy impact on the overall analysis. Specifically, it can have significant impact on the generalization performance of a classification algorithm \cite{Michie95StatLog}, where performance is measured in terms of the ratio of correctly classified instances (i.e., predictive accuracy).

The main tools used for data analysis (e.g., scikit-learn, R, Weka) overlook data pre-processing when it comes to assisting non-expert users in improving the overall performance of their analysis. 
These tools are usually meant for professional users who know exactly which pre-processing operators to apply. However, the staggeringly large number of available pre-processing operators (transformations) overwhelm non-expert users, and they require support.

Hence, our work focuses on assisting the users by reducing the number of pre-processing options to a bunch of potentially relevant ones. The goal is to retain only the transformations that have high positive impact on the analysis. Like this we aim at reducing the time consumed in data pre-processing and at the same time improving the final results of the analysis.
The focus is on classification problems, thus, our method recommends pre-processing operators that improve the performance of a given classification algorithm (e.g., increase the predictive accuracy). 
\newline
\textbf{Contributions.} The main contributions of this paper are as follows:
\begin{itemize}
    \item We apply meta-learning techniques to develop a system that is capable of recommending pre-processing operators (transformations) that positively impact the final result of some classification tasks. 
    Our method is based on training an algorithm to learn the impact of pre-processing operators and then use it to predict and ultimately rank different pre-processing operators.
    
    \item We perform an extensive experimental evaluation to check the accuracy of the rankings with regards to a) the whole set of transformations and b) the top-$K$. For the former, we obtain an accuracy of 63\% as an average for all the algorithms we consider. For the latter (i.e., K=1), the accuracy increases up to 71\% on average.
    
    
    \item We evaluate our rankings with regards to the benefit/gain obtained from the user's point of view and we measure it using a classical information retrieval metric, Discounted Cumulative Gain (DCG). For the whole set of transformations we are as close as 73\% to the gain obtained from the ideal rankings, whereas for the top-1 we are as close as 79\%. 
    

\end{itemize}
The remainder of this paper is organized as follows. In Section 2, we give an overview on data pre-processing and we perform exploratory/empirical analysis on the impact of pre-processing. In Section 3, we discuss meta-learning and its main components. 
In Section 4, we present our tool and proposed method on using meta-learning for data pre-processing. In Section 5, we provide an extensive evaluation of our approach. In Section 6, we discuss the related work and finally, in Section 7, we provide the conclusions and the discussion on our future work.


\section{Data pre-processing}

\begin{table*}[b!]
    \small
    \centering
    \bgroup
    \def\arraystretch{0.8}
    \begin{tabular}{ @{}lcccr@{}} 
        \toprule
        \textbf{Transformation} & \textbf{Technique} & \textbf{Attributes} & \textbf{Input Type} & \textbf{Output Type} \\ \toprule
         Discretization & Supervised & Local & Continuous & Categorical \\ \cmidrule{1-5}
         Discretization & Unsupervised &  Local & Continuous & Categorical \\ \cmidrule{1-5}
         Nominal to Binary & Supervised &  Global & Categorical & Continuous\\ \cmidrule{1-5}
         Nominal to Binary & Unsupervised &  Local & Categorical & Continuous\\ \cmidrule{1-5}
         Normalization & Unsupervised &  Global & Continuous & Continuous \\ \cmidrule{1-5}
         Standardization & Unsupervised &  Global & Continuous & Continuous \\ \cmidrule{1-5}
         Replace Miss. Val. & Unsupervised &  Global & Continuous & Continuous\\ \cmidrule{1-5}
         Replace Miss. Val. & Unsupervised &  Global & Categorical & Categorical\\ \cmidrule{1-5}
         Principal Components & Unsupervised &  Global & Continuous & Continuous \\ 
         \bottomrule
\end{tabular}
\egroup
\caption{List of transformations (data pre-processing operators)}
\label{tbl:transformations}
\end{table*}

Data pre-processing consumes 50-80\% of data analysis time~\cite{MunsonSIGKDD12}. The reason for this is that it encompasses a broad range of activities. Sometimes data needs to be transformed in order to fit the input requirements of the machine learning algorithm (e.g., if the algorithm accepts only data of numeric type, data is transformed accordingly)~\cite{WranglerCHI11}. Sometimes, data requires to be transformed from one representation to another (e.g., from an image (pixel) representation to a matrix (feature) representation)~\cite{DataWranglingEDBT16}, or data may even require to be integrated with other data to be suitable for exploration and analysis~\cite{DataIntegrationPODS02}. Finally, and more importantly, data may need to be transformed with the seldom goal of improving the performance of a machine learning algorithm~\cite{CleaningSIGMOD16}. 
The first two types of transformations are more of a necessity, whereas the latter is more of a choice, and since an abundant number of choices exist, it is time consuming to find the right one.  
In this work, we target the latter type of pre-processing, and as such, the transformations taken into consideration are of the type that can impact the performance of data mining algorithms (i.e., classification algorithms), and they are listed in Table~\ref{tbl:transformations}.  

In Table~\ref{tbl:transformations}, a transformation is described in terms of: 
1) the \textit{Technique} it uses, which can be \texttt{Supervised} --- the algorithm knows the class of each instance and \texttt{Unsupervised} --- the algorithm is not aware of the class, 
2) the \textit{Attributes} it uses, which can be \texttt{Global} --- applied to all compatible attributes, and \texttt{Local} --- applied to specific compatible attributes, 
3) the \textit{Input Type}, which denotes the compatible attribute type for a given transformation, which can be \texttt{Continuous} --- it represents measurements on some continuous scale, or \texttt{Categorical} --- it represents information about some categorical or discrete characteristics, 
4) the \textit{Output Type}, which denotes the type of the attribute after the transformation and it can similarly be \texttt{Continuous} or \texttt{Categorical}.
These operators are the most commonly used ones in the Weka~\cite{HallSIGKDD09} platform, and their implementations are open source\footnote{https://github.com/bnjmn/weka}. 
A short description for each category of transformations from Table~\ref{tbl:transformations} follows.
\\
\noindent\emph{Discretization} - the process of converting or partitioning continuous attributes to discretized or nominal/categorical attributes.\\
\noindent\emph{Nominal to Binary} - the process of converting nominal/categorical attributes into binary numeric attributes.\\
\noindent\emph{Normalization} - the process of normalizing numeric attributes such that their values fall in the range [0,1].\\
\noindent\emph{Standardization} - the process of standardazing numeric attributes so that they have $0$ mean and $1$ variance.\\
\noindent\emph{Missing Value Imputation} - the process of replacing missing values with some other value.\\
\noindent\emph{Principal Component Analysis} - linear dimensionality reduction technique. The goal is to reduce the large number of directly observable features into a smaller set of indirectly observable features.\\
\subsection{Empirical analysis of the overall impact of data pre-processing}
\label{impactOfPP}



\begin{figure*}[!h]
\centering
\includegraphics[width=1.0\textwidth]{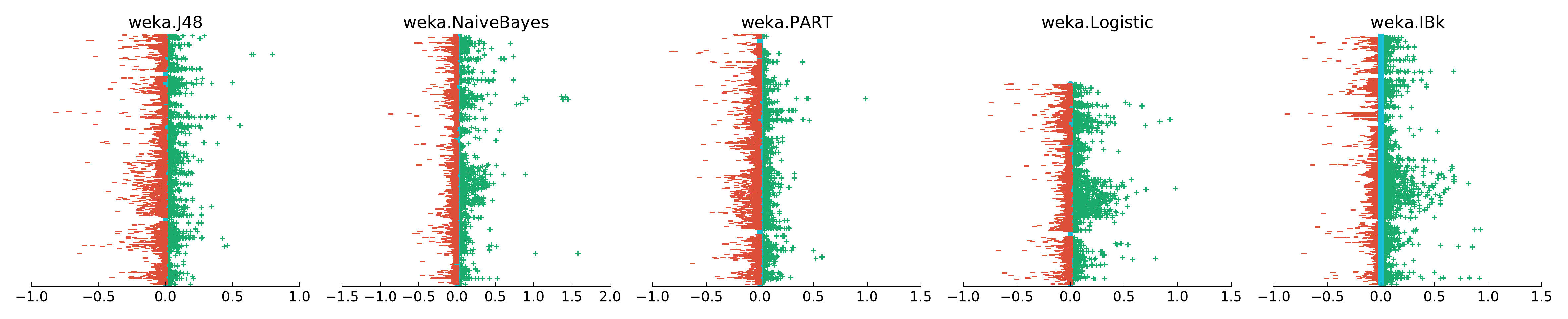}
\caption{Distributions of the (relative) impact induced by transformations in different algorithms}
\label{fig:impact_scatterPlot}
\end{figure*}

Other than theoretical analysis \cite{Kotsiantis06datapreprocessing}, to the best of our knowledge there is not much work on empirically studying the impact of pre-processing operators on real world classification problems. In order to assess the impact of pre-processing, we retrieved 569 datasets from the OpenML~\cite{VanschorenOpenML14} repository and applied the pre-processing operators shown in Table~\ref{tbl:transformations}. We used 5 different classification algorithms (i.e., Nearest Neighbor, Naive Bayes, J48, JRip and Logistic)\footnote{We chose one representative algorithm for 5 different classes of classification algorithms in Weka.} and measured their performance on the datasets before and after the transformations were applied. In Figure~\ref{fig:impact_scatterPlot}, we show the scatter plots of the relative change in predictive accuracy before and after the transformations where applied. In each scatter plot, we visualize the impact of all the transformations applied to datasets for a given algorithm, where green, red, and blue, denote positive, negative, and zero impact, respectively. The total set of transformations applied to all the datasets amounts to 12,000. 

Transformations are applied depending on whether they are Local or Global (as classified in Table~\ref{tbl:transformations}). If a transformation is Global it is applied only once to the set of all compatible attributes (e.g., normalizing all numeric attributes), whereas if it is Local, it is applied to: 1) every compatible attribute separately (e.g., discretizing one attribute at a time), and 2) all the set of compatible attributes (e.g., replacing missing values of all attributes). 

\begin{figure*}[!h]
\includegraphics[width=1.0\textwidth]{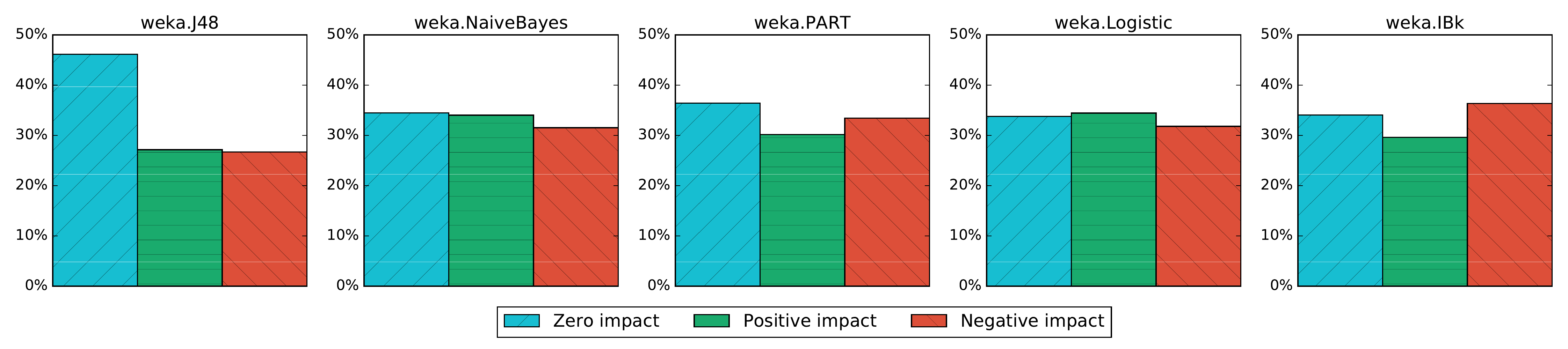}
\caption{The overall impact of transformations expressed in percentage for different algorithms}
\label{fig:impact_barPlot}
\end{figure*}

Observing Figure~\ref{fig:impact_scatterPlot}, one may conclude that: 
\begin{compactitem}
    \item[--] overall, transformations impact the final result of the analysis (i.e., they impact the predictive accuracy of the classification algorithms considered), 
    \item[--] the magnitude of the impact is heterogeneous, and 
    \item[--] there is no clear winner when it comes to the sign of the impact, i.e., transformations do not always impact positively or they do not always impact negatively.
\end{compactitem}
To confirm the latter, in Figure~\ref{fig:impact_barPlot}, we show the percentages of the positive, negative and neutral impacts using bar plots. 
Figure~\ref{fig:impact_barPlot}, shows that transformations are almost uniformly distributed when it comes to the sign of impact, 
which intuitively leads to the conclusion that it may be challenging to distinguish among transformations that affect positively or negatively the final result.

\subsection{Empirical analysis of the impact per pre-processing operator}
\label{sec:bubbles}

\begin{figure*}[!b]
\centering
\includegraphics[width=0.9\textwidth]{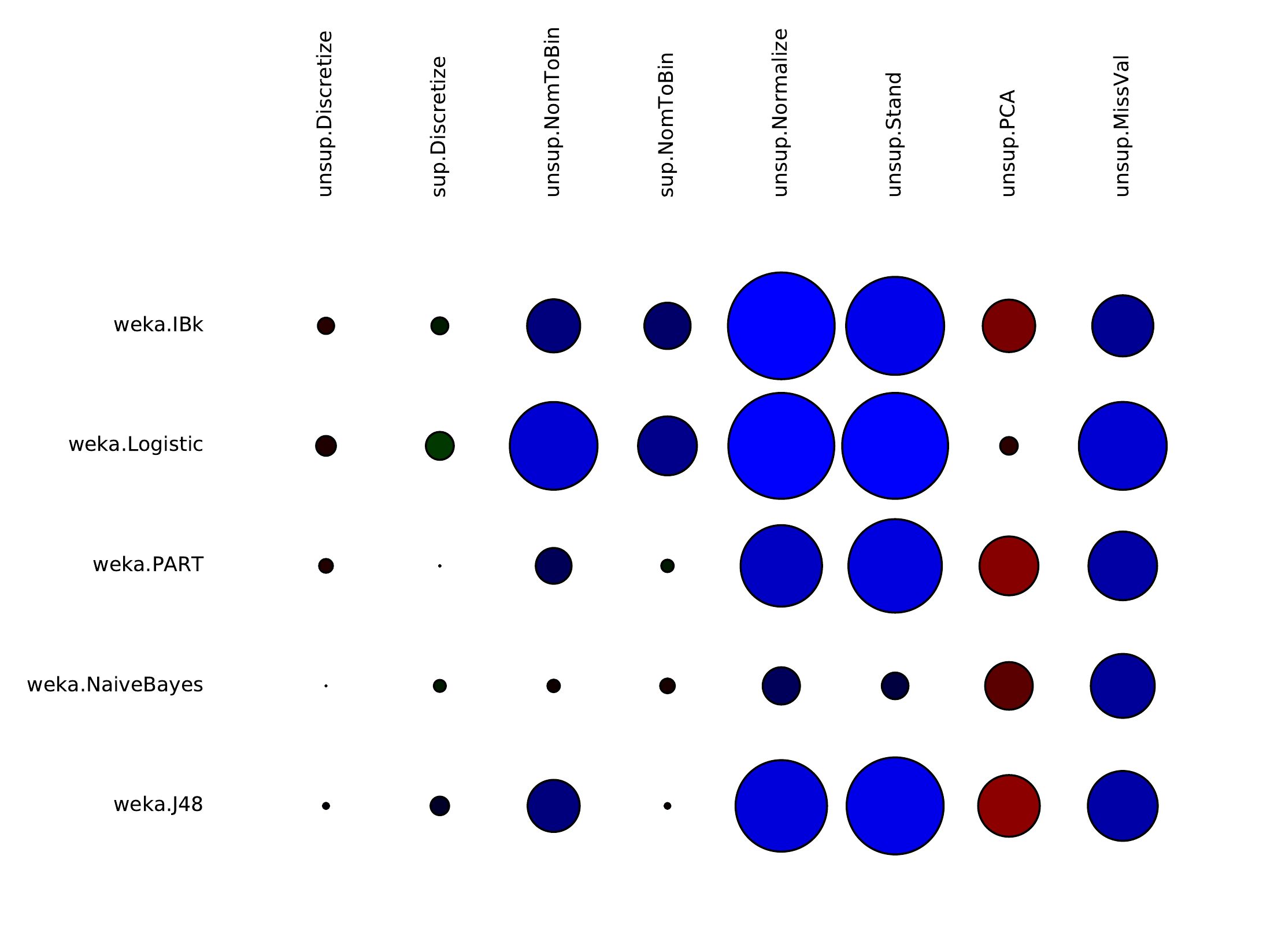}
\caption{Impact of pre-processing operators}
\label{fig:bubbles}
\end{figure*}

In the previous section, we argued that in general, the impact of pre-processing is sound, but it may be difficult to find or predict the transformations that have positive impact. The analysis was performed on all transformations without distinguishing their types. The point now, is to check whether the previous conclusions still hold when we delve into studying categories of transformations separately (e.g., discretization), or conversely to the general picture, there exist some patterns (e.g., discretization has mainly positive impact). 

In Figure~\ref{fig:bubbles}, in a matrix like structure, we show the impact of every transformation from Table~\ref{tbl:transformations}, for every algorithm considered. Circles are sized by the distance from a perfectly uniform distribution of the impact (i.e., 33\% positive, 33\% negative, 33\% neutral), and they are colored by the winner sign (i.e., green if positive, red if negative and blue if neutral impact is the winner). Thus, the bigger the circle and the sharper the color, the more obvious the pattern for a given transformation. 

For instance, suppose Nearest Neighbor (IBk) for Normalization has a distribution of $10\%,10\%$, $80\%$ for positive, zero and negative impact, respectively, and NaiveBayes for the same transformation has a distribution of $45\%,25\%,30\%$. Then, the sizes of the circles are determined by the euclidean distance between $(10,10,80)$ and $(33,33,33)$, for the first algorithm, and the distance between $(45,25,30)$ and $(33,33,33)$, for the second algorithm.  The distance for the first algorithm $(57.15)$ is obviously higher than the distance for the second algorithm $(14.73)$, and hence the size of the circle. 
Furthermore, to define the colors of the circles, the distributions for positive, negative and zero impact, participate proportionally to the RGB (red, green, blue) coloring scheme. Hence, for the above mentioned example for the first algorithm color blue will be more decisive, and for the second green. Yet, for the first algorithm, the color will be sharper than for the second, because of the values being higher.

The patterns emerging from the plot may help us in two directions. First, they can be used to devise basic rules or heuristics, i.e., if a transformation has a big blue circle for a given algorithm, then the transformation can be discarded because it is basically of no use for that particular algorithm, since most of the time it does not affect the final result. Secondly, they enable us to determine the more difficult transformations in terms of finding the impact to the final analysis, i.e., if a transformation has a small circle for a particular algorithm, it means that the distribution of the impact is close to uniform and hence a simple rule may not help in finding the impact of the transformation. The latter rises the need for developing more sophisticated techniques for discovering the impact of transformations. To this end, we propose to learn the impact of transformations using meta-learning, and we delve into more details of this, in the next section.

\subsubsection{Simple heuristics/rules as result of the empirical study}
\label{sec:simple-heuristics}
In Figure~\ref{fig:bubbles}, circles of bigger size give clear patterns for devising simple heuristics. Notice that the bigger circles are usually of blue color. This means that the transformations for the corresponding algorithms are not of much use, since they do not impact the performance of the algorithms on the tested datasets. Some of the blue circles are obviously expected. For instance, it is well known that Normalization and Standardization do not impact the performance of Decision Trees (i.e., J48). Hence, a simple heuristic would be that, when a Decision Tree is chosen, Normalization and Standardization should not appear in the palette of possible transformations. The same holds for Logistic Regression. These transformations do not impact its performance. However, a counter-intuitive pattern appearing, is that of Normalization and Standardization with Nearest Neighbor. One would expect an impact of transformations, yet the circles are big and the color is blue, implying no impact. 

We studied the internals of the Weka implementation in order to understand why was this happening. It resulted that Nearest Neighbor in Weka, internally uses a normalized Distance algorithm. Hence, the Normalization/Standardization is implicitly performed inside the learning algorithm and as a consequence, an external Normalization does not impact the performance. 

The rest of the transformations have smaller circles in Figure~\ref{fig:bubbles}, which implies that they have impact on the performance of algorithms. However, in the case of PCA, for instance, although smaller, the circles are reddish. This indicates that although PCA impacts the performance, it needs to be carefully performed. A default or careless application of PCA may lead to a negative impact on the overall performance, which indicates that PCA should be used by experts. However, it is still interesting to discover the transformations that have negative impact and advise the non-expert user to avoid such transformations.


\section{Meta-learning for predicting the impact of pre-processing}

\label{sec:meta-learning}
\begin{figure*}[!b]
\centering
\includegraphics[width=0.7\textwidth]{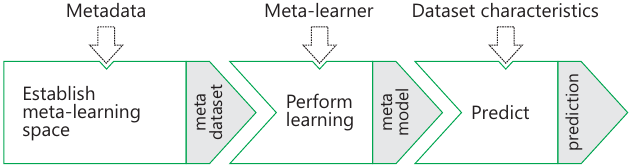}
\caption{Meta-learning process}
\label{fig:ML}
\end{figure*}

Meta-learning is a general process used for predicting the performance of an algorithm on a given dataset. It is a method that aims at finding relationships between dataset characteristics and data mining algorithms \cite{BrazdilBook08}. Given the characteristics of a dataset, a predictive meta-model can be used to foresee the performance of a given data mining algorithm. For instance, in a classification problem, meta-learning can be used to predict the predictive accuracy of a classification algorithm on a given dataset and hence provide user support in the mining step.

Meta-learning can also be used to provide support in the pre-processing step~\cite{BilalliCSI17,Gemp17AAAI}. 
This can be done by learning the impact of data pre-processing operators (transformations) on the final result of the analysis. That is to say, detecting the transformations that have the most positive impact on the final analysis (i.e., transformations that after being applied on a dataset of classification type, increase the accuracy of a given classification algorithm on that dataset). This way, meta-learning pushes the user support to the data pre-processing step by enabling a ranking of transformations according to their relevance to the analysis.
The ranking is made possible through the following three phases, shown in Figure~\ref{fig:ML}. 

First, a meta-learning space is established using metadata. The metadata consists of dataset characteristics along with some performance measures for data mining algorithms on those particular datasets. Then, the meta-learning phase generates a model (i.e., predictive meta-model), which defines the area of competence of the data mining algorithm~\cite{KalousisNoemon01}. Finally, when a transformed dataset (i.e., a transformation was applied on the dataset) arrives, the dataset characteristics are extracted and fed into the predictive meta-model, which predicts the performance of the algorithm on the transformed version of the dataset. At this point, we are able to obtain predictions for different transformed datasets (e.g., different transformations applied to the same dataset). By comparing the obtained predictions for the different transformations, we are able to rank the transformations depending on their predicted impact on the given dataset. This concludes the prediction phase.

To enable meta-learning, the first thing to do for each classification algorithm is to create a dataset, that is, a matrix-like structure consisting of variables/features (predictors) and a response. The variables of this dataset are dataset characteristics -- in this case, characteristics of the transformed datasets (e.g., number of instances, number of missing values, etc.). The response is a performance metric of the classification algorithm (e.g., predictive accuracy). Since all these are metadata, this dataset is called a \textbf{meta-dataset}. Consequently its variables are referred to as \textbf{meta-features} and the response variable is named as \textbf{meta-response}. Furthermore, the process of learning on top of this meta-dataset is referred to as \textbf{meta-learning} and the learning algorithm used, is referred to as \textbf{meta-learner}.
The meta-features, the meta-response and the meta-learner are the key ingredients, and we will delve into details of each one of them, in the following sections.

\subsection{Meta-features}

\begin{table*}[!t]
    \small
    \centering
    \begin{tabular}{@{}lllc@{}}
    \toprule
        \textbf{No} & \textbf{Name} & \textbf{Type}  & \textbf{Modifiable} \\ \toprule
        1..2 & [Number$|$Percentage] of Continuous Attributes & Continuous & Yes \\  \cmidrule{1-4}
        3..6 & Min[Means$|$Std$|$Kurtosis$|$Skewness] of Continuous Attributes & Continuous & Yes \\ \cmidrule{1-4}
        7..10 & Mean[Means$|$Std$|$Kurtosis$|$Skewness] of Continuous Attributes & Continuous & Yes \\ \cmidrule{1-4}
        11..14 & Max[Means$|$Std$|$Kurtosis$|$Skewness] of Continuous Attributes & Continuous & Yes \\ \cmidrule{1-4}
        15..17 & Quartile [1$|$2$|$3] of Means of Continuous Attributes & Continuous & Yes \\ \cmidrule{1-4}
        18..20 & Quartile [1$|$2$|$3] of Std of Continuous Attributes & Continuous  & Yes \\ \cmidrule{1-4}
        21..23 & Quartile [1$|$2$|$3] of Kurtosis of Continuous Attributes & Continuous & Yes \\ \cmidrule{1-4}
        24..26 & Quartile [1$|$2$|$3] of Skewness of Continuous Attributes & Continuous & Yes \\ \cmidrule{1-4}
        27 & Number of Categorical Attributes & Categorical & Yes \\ \cmidrule{1-4}
        28 & Number of Binary Attributes & Categorical & Yes \\ \cmidrule{1-4}
        29 & Percentage of Categorical Attributes & Categorical & Yes \\ \cmidrule{1-4}
        30 & Percentage of Binary Attributes & Categorical & Yes \\ \cmidrule{1-4}
        31..33 & [Min$|$Mean$|$Max] Attribute Entropy & Categorical & Yes \\ \cmidrule{1-4}
        34..36 & Quartile [1$|$2$|$3] Attribute Entropy & Categorical & Yes \\ \cmidrule{1-4}
        37..39 & [Min$|$Mean$|$Max] Mutual Information & Categorical & Yes \\ \cmidrule{1-4}
        40..42 & Quartile [1$|$2$|$3] Mutual Information & Categorical & Yes \\ \cmidrule{1-4}
        43 & Equivalent Number of Attributes & Categorical & Yes \\ \cmidrule{1-4}
        44 & Noise to Signal Ratio & Categorical & Yes \\ \cmidrule{1-4}
        45..48 & [Min$|$Mean$|$Max$|$Std] Attribute Distinct Values & Categorical & Yes \\ \cmidrule{1-4}
        49 & Number of Instances & Generic & Yes \\ \cmidrule{1-4}
        50 & Number of Attributes & Generic & Yes \\ \cmidrule{1-4}
        51 & Dimensionality & Generic & Yes \\ \cmidrule{1-4}
        52,53 & [Number$|$Percentage] of Missing Values & Generic & Yes \\ \cmidrule{1-4}
        54,55 & [Number$|$Percentage] of Instances with Missing Values & Generic & Yes \\ \cmidrule{1-4}

        56 & Number of Classes & Generic  & No \\ \cmidrule{1-4}
        57 & Class Entropy & Generic & No \\ \cmidrule{1-4}
        58,59 & [Minority$|$Majority] Class Size & Generic & No \\ \cmidrule{1-4}
        60,61 & [Minority$|$Majority] Class Percentage & Generic & No \\ \bottomrule
    \end{tabular}
    \caption{Meta-features (Dataset characteristics)}
    \label{tbl:metadata}
\end{table*}

Meta-features characterize a dataset, and two main classes have been proposed:
\begin{itemize}
\item \emph{General measures:} include general information related to the dataset at hand. To a certain extent they are conceived to measure the complexity of the underlying problem. Some of them are: the number of instances, number of attributes, dataset dimensionality, ratio of missing values, etc.
\item \emph{Statistical and information-theoretic measures:} describe attribute statistics and class distributions of a dataset sample. They include different summary statistics per attribute like mean, standard deviation, class entropy, etc.
\end{itemize}

Additional meta-features measuring the association between the predictors and the response have been proposed. These measures are grouped into the \textit{Landmarking and Model-based} class \cite{Pfahringer2000,Peng2002}. This class is related to measures asserted with simple machine learning algorithms, so called \textit{landmarkers}, and their derivatives based on the learned models. They include error rates and area under the roc curve (AUC) values obtained by landmarkers such as 1NN, DecisionStump or NaiveBayes.
When performed on bigger datasets, however, these simple machine learning algorithms (landmarkers) introduce significant computational costs~\cite{BilalliAMCS17}. Hence, we do not consider them as dataset characteristics and they do not participate as meta-features in our experiments.

The meta-features we specifically consider are shown in Table~\ref{tbl:metadata}. These are the set of meta-features extracted from OpenML~\cite{VanschorenOpenML14}. OpenML is an open science platform developed with the aim of allowing researchers to share their datasets, implementations, and experiments (machine learning and data mining) in a way that they can easily be found and reused by others. OpenML is the biggest source of data and metadata for advancing meta-learning studies.  

In Table~\ref{tbl:metadata}, column \textit{Type}, specifies the type of the meta-feature, and it can be \texttt{Continuous} -- the meta-feature can be extracted only from datasets that contain attributes of continuous type, \texttt{Categorical} -- the meta-feature can be extracted only from datasets that contain attributes of categorical type, \texttt{Generic} -- the meta-feature can be extracted from any dataset, regardless of the types of its attributes.

Column \textit{Modifiable} indicates whether the meta-features are modifiable through the transformations we use (listed in Table~\ref{tbl:transformations}). If meta-features are not modifiable/transformable, they are not considered, because they remain constant and they do not reflect the impact of transformations.

Note that the ultimate goal is to predict the impact of transformations, and the impact per se, is measured as the relative difference of the performance of the algorithm before and after the transformation was applied. To this end, to the set of meta-features we consider, we attach also the base performance of the classification algorithm (i.e., the performance before the transformation is applied) and in addition we add features that capture the difference between the meta-features before and after the transformation was applied. We call these features \textbf{delta meta-features}. As a result, every meta-feature has its corresponding delta meta-feature. For instance, let us say that in a given dataset, before applying a transformation, the \textit{number of continuous attributes} is $5$. Assume we apply a transformation that is discretizing only one continuous attribute, then, the number of continuous attributes becomes $4$ and thus the delta of this feature is $-1$ (i.e., the \textit{delta of the number of continuous attributes}).

Taking the deltas into account the total set of meta-features becomes large. We apply meta-feature extraction and selection in order to select only the most informative (with more predictive power) meta-features. Details on the meta-feature extraction/selection performed can be found in our previous work \cite{BilalliAMCS17}.

\subsection{Meta-response}

The goal of meta-learning is to correctly predict the impact of transformations on the performance of machine learning algorithms. Different measures can be used to evaluate the performance of machine learning algorithms. Since we are dealing with classification problems, and hence the algorithms we consider are of classification type, the performance is usually measured in terms of \textit{predictive accuracy}, \textit{precision}, \textit{recall}, or \textit{AUC}~\cite{Hand09ML}. 
Moreover, classification algorithms are usually evaluated using either 10-fold cross-validation or leave-one-out validation (LOOV)~\cite{Kohavi95}. 
In Table \ref{tbl:performance-measures}, formulas for calculating these measures are given.
Briefly, \textit{Accuracy} is a measure of the overall effectiveness of a classifier. \textit{Precision} is the class agreement of the instance labels with the positive labels given by the classifier. \textit{Recall} measures the effectiveness of a classifier to identify positive labels. Finally, one can think of \textit{AUC} as the classifier's ability to avoid false classification. 
For more details regarding these measures and how they extend to multi-class classification problems, we refer the reader to \cite{Sokolova09:IPM}.

These measures are collected before and after the transformations have been applied. The relative difference between the performance obtained after the transformation and the base performance (performance obtained before the transformation) is the impact of a transformation on the predictive power of a classification algorithm, and this is the meta-response. Based on the meta-features and delta meta-features mentioned previously, the goal of the meta-learner is to correctly predict this impact, which can be positive -- if the transformation helps on improving the performance, negative -- if the performance decreases after the transformation, and zero -- if the performance remains the same.

\begin{table}[h!]
    \small
    \centering
    \begin{threeparttable}
    \begin{tabular}{@{}lc@{}}
    \toprule
    \textbf{Measure} & \textbf{Formula} \\ \midrule
    Accuracy &\parbox{5cm}{
    \begin{equation*}
        \frac{TP+TN}{TP+FP+FN+TN}
    \end{equation*}
    } \\ \cmidrule{1-2}
    Precision & \parbox{5cm}{
    \begin{equation*}{TP}/{(TP+FP)}
    \end{equation*}
    } \\ \cmidrule{1-2}
    Recall & \parbox{5cm}{
    \begin{equation*}{TP}/{(TP+FN)}
    \end{equation*}
    } \\ \cmidrule{1-2}
    AUC & \parbox{5cm}{
    \begin{equation*} Prob(X2>X1)\end{equation*}}
    \\ \bottomrule
    \end{tabular}
    \begin{tablenotes}
    \footnotesize
    \item TN - True Negatives; TP - True Positives; FN - False Negatives; FP - False Positives; X1, X2 - Score functions of the classes 
    \end{tablenotes}
    \end{threeparttable}
    \caption{Performance evaluation measures for classification algorithms}
    \label{tbl:performance-measures}
\end{table}

\subsection{Meta-learner}
\label{sec:meta-learner}
Given that meta-features (including delta meta-features) and meta-response candidates are defined, the next step is to define the meta-learning problem. Since the meta-response -- the impact of transformations, is of continuous (numeric) type, the learning problem naturally fits to a regression type. Yet, we are interested in finding the transformations that either positively or negatively impact the data analysis, without necessarily -- though preferably, needing to know the exact amount of impact. As a matter of fact, the problem may as well be defined as a classification problem, where three classes would be required, positive, negative, and zero.     
Regardless the type of the learning chosen, formally, the problem can be defined as follows. 
Given algorithm $A$ and a limited number of training data $D={(\textbf{x}_1,y_1)...(\textbf{x}_n,y_n)}$, the goal is to find a meta-learner with optimal/good generalization performance. Generalization performance is estimated by LOOV, which splits the training data into $n$ partitions $D^{(1)}_{valid},...,D^{(n)}_{valid}$, and sets $D^{(i)}_{train}=D\backslash D^{(i)}_{valid}$ for $i=1,...,n$. Note that $\textbf{x}\in{x_1,x_2 ...x_n}$ are the meta-features and delta meta-features and $y_i$ is the impact of the transformation on the performance of algorithm $A$ run on that particular transformed dataset. Hence, $\textbf{x}$ and $y$ altogether are the extracted metadata. Since $y$ consists of 4 different performance measures (shown in Table~\ref{tbl:performance-measures}) for algorithm runs, we build meta-datasets for each specific measure separately (we discuss only the results on predictive accuracy). Then, for each meta-dataset, we generate meta-models -- using a meta-learner. 

\section{PRESISTANT}
\begin{figure*}[!t]
\centering
\includegraphics[width=0.7\textwidth]{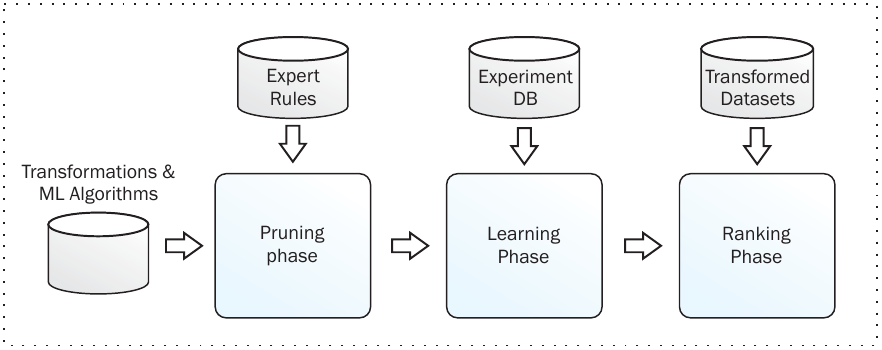}
\caption{Overview of PRESISTANT}
\label{fig:system-overview}
\end{figure*}

When dealing with a classification problem, the non-expert data analyst 
has to choose from a large number of machine learning algorithms, and in addition, he/she encounters a plethora of different pre-processing options.
Once the classification algorithm is chosen (i.e., one of the algorithms considered), PRESISTANT assists the user by reducing the number of pre-processing options to only a set of relevant ones (i.e., operators that have positive impact).
To do this, PRESISTANT uses a method consisting of three phases, shown in Figure~\ref{fig:system-overview}. 

In the first phase, rules are applied to prune transformations such that the search space is reduced. In the second phase, a model is trained to learn the impact of transformations on the performance of  classification algorithms. Finally, in the third phase, the trained model is used to rank the newly arriving  transformed versions of datasets. 

\subsection{Pruning phase}

Given that there is an overwhelming number of different transformations that can be applied to a dataset, in Section~\ref{sec:simple-heuristics}, we argued that simple rules can help on discarding transformations that have no impact. This translates to having a repository of Expert Rules (see Figure~\ref{fig:system-overview}), that can be extended to contain any types of rules that may be known in advance and that reduce the number of potential transformations to be applied on datasets. 
Our first basic set of rules is derived from the experiments whose results were shown in Figure~\ref{fig:bubbles}, where we for instance define rules in order to exclude Standardization and Normalization when considering algorithms like, IBk, Logistic, and J48.

\subsection{Learning phase}
Two important activities are performed in the learning phase. First, a meta-database (i.e., set of meta-datasets) is generated for all the classification algorithms considered (see Algorithm~\ref{alg:create_meta-database}), and then on top of it, a learning algorithm is applied (see Algorithm~\ref{alg:create_models}). As a result, a statistical model (meta-model) is generated for every classification algorithm considered. 

The inputs required to construct the meta-database, are,  datasets, transformations -- that are likely to improve the performance of classification algorithms, and the classification algorithms in consideration.

For the sake of simplicity, let us consider that we want to create the meta-dataset for a single classification algorithm. In line 5 of Algorithm~\ref{alg:create_meta-database}, we first extract the dataset characteristics (i.e., meta-features from the original non-transformed datasets). 
Next, we apply all the available transformations to all the datasets and hence obtain transformed datasets, see line 8. We extract the meta-features from the transformed datasets in line 9, and take the difference between them and the meta-features from the original non-transformed datasets in line 10. Like this, we obtain the delta meta-features. Furthermore, to both original non-transformed datasets -- line 6, and the transformed ones -- line 11, we apply the classification algorithm and then take the relative difference between their corresponding performance measures (e.g., predictive accuracy) -- line 12. The latter is the meta-response, which together with the meta-features of the non-transformed version of the dataset, the delta meta-features, and the performance measure of the original dataset compile the complete set of metadata -- see line 13.

\begin{algorithm*}[!h]
\textbf{Output:} $meta\_db[1..\#algs][1..\#trans][1..\#metadata];$ \Comment{meta-database; meta-dataset per classification alg.}
\caption{Establish meta-database}
\label{alg:create_meta-database}
\begin{algorithmic}[1]
\Function{CreateMetaDB}{$datasets[]$,$transformations[]$,$classificationAlgs[]$}
    \State $metadata[] = \varnothing;$ \Comment{the set of metadata to be collected}
    \For {each algorithm $alg$ in $classificationAlgs$}
        \For {each dataset $ds$ in $datasets$}
            \State $ds\_mf[] =$ \textproc{ComputeMetaFeatures}($ds$); \Comment{see Table~\ref{tbl:metadata}}
            \State $ds\_pm=$ \textproc{GetPerformanceWith10FoldCV}($alg$,$ds$); \Comment{see Table~\ref{tbl:performance-measures}}
            \For {each transformation $tr$ in $transformations$}
                \State $trans\_ds =$ \textproc{ApplyTransformation}($tr$,$ds$);
                \State $trans\_ds\_mf[] =$ \textproc{ComputeMetaFeatures}($trans.ds$);
                \State $\Delta mf[] = trans\_ds\_mf[] - ds\_mf[];$
                \State $trans\_ds\_pm =$ \textproc{GetPerformanceWith10FoldCV}($alg$,$ds$);
                \State $mr = {|trans\_ds\_mr - ds\_pm|}\mathbin{/}{ds\_mr};$ \Comment{relative diff., meta-response}
                \State $metadata[] = trans\_ds\_mf[] \cup \Delta mf[] \cup ds\_pm \cup mr;$
                \State $meta\_db[alg][trans\_ds] = metadata[];$
            \EndFor
        \EndFor
	\EndFor
	\State \Return $meta\_db;$
\EndFunction
\end{algorithmic}
\end{algorithm*}

Once a meta-dataset for each classification algorithm is obtained, next a learning algorithm (i.e., meta-learner) is applied on top -- line 6 of Algorithm~\ref{alg:create_models}, and as a result, a meta-model (i.e., statistical model) for each of the classification algorithms is obtained. PRESISTANT uses the Random Forest~\cite{Breiman01RandomForest} algorithm as meta-learner. The XGBoost~\cite{XGBoost16Chen} algorithm was also tested as meta-learner. Similar results were obtained, yet, Random Forest is easier to interpret.

\begin{algorithm*}[!t]
\textbf{Input:} $datasets[..]$ \Comment{available datasets of classification type} \\
$transformations[..]$ \Comment{set of transformations to be applied} \\
$classificationAlgs[..]$ \Comment{available classification algorithms} \\
\textbf{Output:} $models[]$ \Comment{meta-model for each algorithm}
\caption{Create meta-models}
\label{alg:create_models}
\begin{algorithmic}[1]
\Function{PerformMetaLearning}{}
    \State $meta\_db = $\textproc{CreateMetaDB}($datasets[]$,$transformations[]$,$classificationAlgs[]$);
    \State $meta\_learner = RandomForest()$; \Comment{a meta learner of choice}
    \State $models[] = \varnothing$;
    \For {each algorithm $alg$ in $classificationAlgs$}
        \State $models[alg] =$ \textproc{ApplyMetaLearner}($meta\_learner$,$meta\_db[alg][][]$); 
    \EndFor
    \State \Return $models;$
\EndFunction
\end{algorithmic}
\end{algorithm*}

\begin{algorithm*}[!b]
\textbf{Input:} $models[..];$ \Comment{meta-model for each algorithm} \\
$transformations[..];$ \Comment{set of transformations to be applied} \\
$ds$ \Comment{new dataset chosen by the user} \\
\textbf{Output:} $transformations[..];$ \Comment{trans. ordered according to predicted impact}
\caption{Recommend transformations}
\label{alg:recommending_phase}
\begin{algorithmic}[1]
\Function{RankTransformations}{$datasets[]$,$transformations[]$,$classificationAlgs[]$}
    \State $predictions[][];$ \Comment{predictions for transformed datasets}
    \State $ds\_mf[]=$\textproc{ComputeMetaFeatures}($ds$);
    \State $ds\_pm=$\textproc{GetPerformanceWith10FoldCV}($alg$,$ds$);
    \For {each transformation $tr$ in $transformations$}
        \State $trans\_ds=$ \textproc{ApplyTransformation} ($tr$,$ds$);
        \State $trans\_ds\_mf =$\textproc{ComputeMetaFeatures}($ds$);
        \State $\Delta mf[] = trans\_ds\_mf[] - ds\_mf[];$
        \State $features[] = trans\_ds\_mf[] \cup \Delta mf[] \cup ds\_pm;$
        \State $predictions[tr]=$ \textproc{ApplyModel}($features[]$,$models[classAlg]$); \Comment{predict perf.}
    \EndFor
     \State $transformations=$\textproc{RankByProbabilities}($predictions$,$desc=true$);
     \State \Return $transformations;$
\EndFunction
\end{algorithmic}
\end{algorithm*}

\subsection{Ranking/Recommending phase}
The recommending phase starts when a user wants to analyze a dataset. He/she selects an algorithm to be used for the analysis and the system automatically recommends transformations to be applied, such that the final result is improved. This phase is described in Algorithm~\ref{alg:recommending_phase}. In Algorithm~\ref{alg:recommending_phase}, first the meta-features and the performance of the classification algorithm are extracted from the original non-transformed dataset in lines 4 and 5, respectively. Next, different transformations are applied to the dataset and from each transformed version of the dataset the necessary features (i.e., meta-features, delta meta-features) are computed -- see lines 6-10. The extracted features are then fed to the predictor in line 11. The predictor in line 11, does no more than applying an already existing meta-model to the extracted features, to find the predicted impact of a transformation on the performance of the algorithm. 

After the predicted impacts are obtained for all the transformations, they are ranked in descending order -- using the probabilities of being positive; provided by the model, in line 12.

Although it may seem computationally costly to execute transformations and then predict the performance of algorithms on the transformed datasets, notice that this is orders of magnitude less costly than applying the classification algorithm after each transformation. This in fact, is the strongest point of our approach.



\section{Evaluation}

We perform an experimental study considering a set of classification algorithms and hundreds of datasets. 
The aim of the experiments is two-fold. First, to asses the performance of the method in terms of the quality/effectiveness of the recommendations from the meta-learner perspective, that is, we try to answer the question ``How good are the predictions?". Second, to assess the value of the recommendations from the user perspective, we try to answer the question ``How valuable/profitable are the recommendations?" 

To enable the use of the entire set of datasets in the experiments, we use the leave-one-out validation (LOOV) method. This entails that for each classification algorithm considered, when building the meta-models, if a dataset is used in testing, the same is not considered in the training.

We performed experiments treating the problem as classification and as regression. Similar results were obtained in both cases. We discuss only the results obtained with classification.

\subsection{Evaluation of the quality of predictions}

Predictions provided by the meta-model enable the ranking of transformations. The list of recommended transformations can be very large in case a lot of transformations are considered (e.g, with different parametrization). 
 One may be interested in the whole set of transformations (e.g., ``recommending all good items" in collaborative filtering), or only on the top-K transformations, K being an arbitrary number (e.g., ``recommending some good items" in collaborative filtering). The latter is more realistic since the greater the ranked position, the less valuable a transformation is for the user, because the less likely it is that the user will examine the transformation due to time, effort and cumulated information from transformations already seen/applied~\cite{IRevalSIFIR00}.
We performed evaluations both considering the whole set of transformations and considering only the top-K.

\subsubsection{Evaluation of the quality of the whole set of transformations}
When treated as classification, the problem translates into a multi-class classification problem with three classes (i.e., positive, negative, zero) in the response variable. Given that it is a multi-class problem and knowing that all the classes do not have the same importance, we cannot use the traditional binomial confusion matrix for the evaluation. For instance, regarding the importance of classes, miss-predicting a zero transformation does not have the same impact as miss-predicting a positive or negative transformation. 

\begin{table}[!t]
\centering
\includegraphics[width=0.4\textwidth]{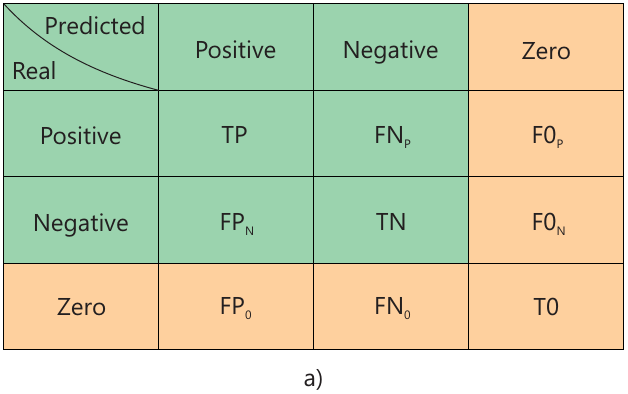}
\includegraphics[width=0.4\textwidth]{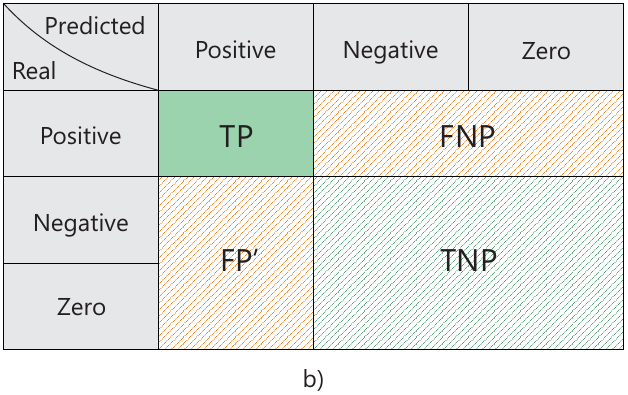}
\caption{Confusion matrices}
\label{fig:confusion-matrix_classification}
\end{table}

Therefore, in Table~\ref{fig:confusion-matrix_classification}a, we devise a confusion matrix that consists of two parts. The inner (green) part is the traditional confusion matrix for the positive and negative predictions and the outer (orange) part, is the one that takes into account the zero class.

Furthermore, since datasets have varying numbers of attributes, they do not have the same number of transformations applied to them i.e., some datasets can have more transformations than others (see Table~\ref{tbl:transformations}). To give equal importance to every dataset, regardless of the number of transformations, we assign weights to their corresponding transformations (i.e., transformed versions of the same dataset): $w_{T_d} = 1/|T_d|$, where $|T_d|$ is the total number of transformations applied to dataset $d$. Like this, each dataset has weight equal to $1$.

Using the matrix shown in Table~\ref{fig:confusion-matrix_classification}a, per dataset $d$ and for the set of its transformations $T_d$ with their corresponding weights $w_{T_d}$, we evaluate our system calculating the Predictive accuracy ($PA_d$), Precision ($Pr_d$), Overall recall ($OR_d$) and G-measure ($G_d$), defined as follows:

\begin{align*}
&PA_d = \frac{TP+TN}{TP+FN_P+FP_N+TN}\ \ 
Pr_d = \bigg(\frac{TP}{TP+FP_N}+\frac{TN}{TN+FN_P}\bigg)/2 \\ \\
&OR_d = \frac{TP+FN_P+FP_N+TN}{TP+FN_P+FP_N+TN+(F0_P+F0_N)}\ \
G_d = 2\bigg(\frac{PA_d\times OR_d}{PA_d+OR_d}\bigg)\\ 
\end{align*}
where $TP$ represents the number of true positives, $FN_P$ the number of false negatives, $FP_N$ the number of false positives, and $TN$ the number of true negatives. 
Furthermore, $F0_P$ represents the number of transformations that are predicted as zero, but in reality they are positive, $F0_N$ represents the zero predicted transformations that are in reality negative. 
Finally, $FP_0$ are positive predictions that in reality have zero impact, $FN_0$ negative predictions that in reality have zero impact, and $T0$ are the true zeros. Notice, that $FP_0$ and $FN_0$ are less harmful than $F0_P$ and $F0_N$, since predicting a transformation as zero and then having a positive impact in real ($F0_P$), is worse than predicting a transformation as positive and then having zero impact in real ($FP_0$). The same applies for $FN_0$ when compared to $F0_N$. 

\begin{figure*}[!t]
\centering
\includegraphics[width=0.8\textwidth]{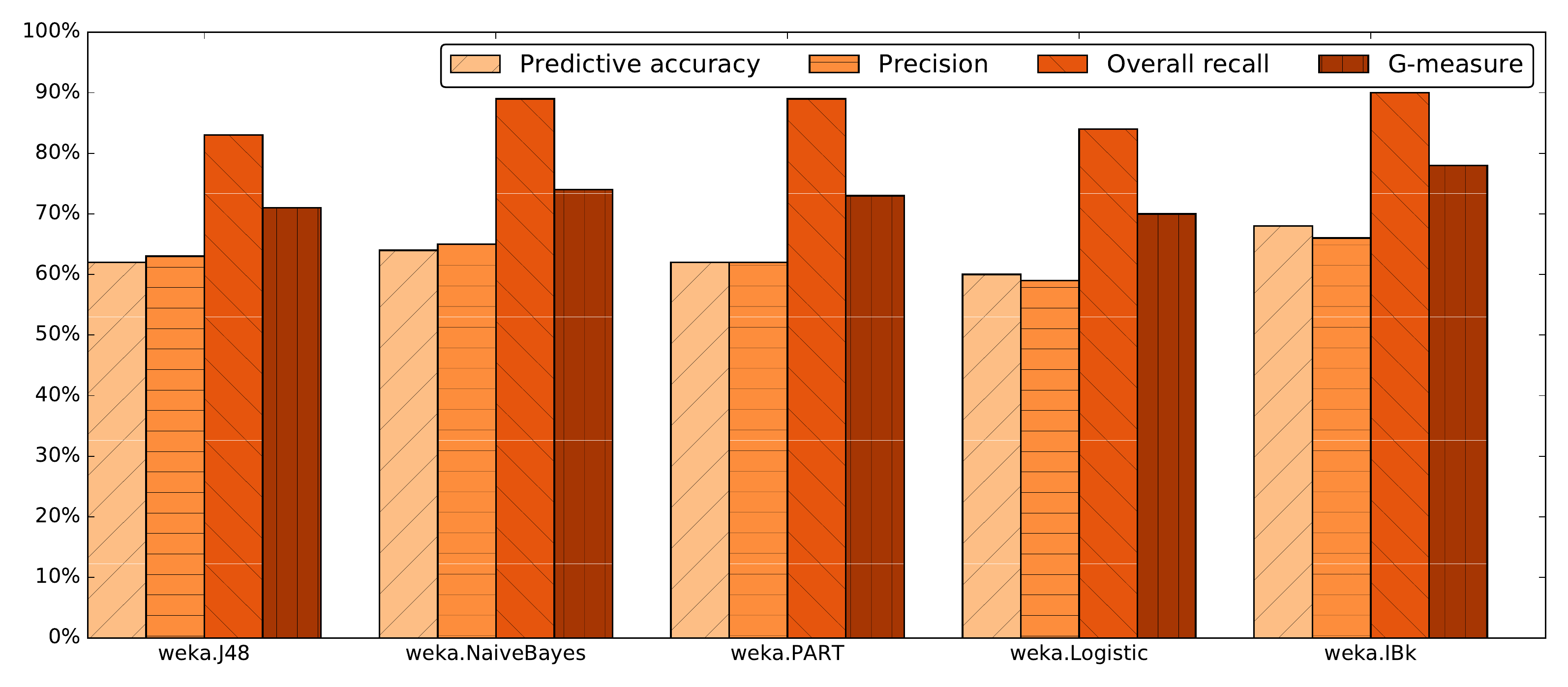}
\caption{Results obtained when evaluating the whole set of transformations}
\label{results:all_transformations}
\end{figure*}

The aforementioned measures are calculated for individual datasets $d$. Averaging the individual measures over all datasets with at  least  one  relevant  transformation,  we  obtain  the  mean  Predictive accuracy $PA$, Precision $Pr$, Overall recall $OR$, and G-measure $G$.
In the experiments performed on 569 datasets with LOOV, we obtained the results shown in Figure~\ref{results:all_transformations}.

The results show that, on average, if a user selects any transformation from the whole list of possible transformations ($T_d$), the system provides an accuracy of 63\%.

\subsubsection{Evaluation of the quality of the top-K recommendations}

Since real users are usually concerned only with the top part of the recommendation
list, a more practical approach is to consider the number of a datasets' relevant transformations ranked in the top-K positions. 

Many details need to be considered in order to perform a proper evaluation of the K positions.

First of all, for the sake of simplicity, let us use the confusion matrix shown in Table~\ref{fig:confusion-matrix_classification}b, where we denote as True Non Positive ($TNP=TN+F0_N+FN_0+T0$), a transformation that is predicted as non-positive and it is non-positive in real (i.e, after executing the classification algorithm the transformation has no positive impact). False Non Positive ($FNP=FN_P+F0_P$), a transformation that is predicted as non-positive but is positive in real. Finally, $FP'$, a transformation that is predicted as positive but in reality can have either negative or zero impact.

Next, notice that for each dataset, there are $L$ transformations that have real positive impact, and the system (in practice) recommends $y$ transformations that are predicted to have a positive impact. 


To be able to perform evaluations for all the datasets, including the datasets with L=0 (i.e., datasets that do not have any transformations that have real positive impact), and to be able to calculate average measures for datasets with different L, for all the positions in the ranking, we rank the transformations as follows:
first we rank the $y$ positively predicted transformations by their probability of being positive (the highest goes first). Next, we append the remaining real positive transformations (if any are left) up to $L$. Finally, we append all the remaining transformations ranked by their probability of being positive.

This ranking allows us to perform evaluations for each position $K$ for any dataset with $L$ real positive transformations. The results of the evaluations form a matrix of size $[L,K]$, 
and the possible evaluation we can have is shown in Table~\ref{evaluation:our}, where two possible scenarios are considered,

\begin{table}[!b]
\centering
\includegraphics[width=0.5\textwidth]{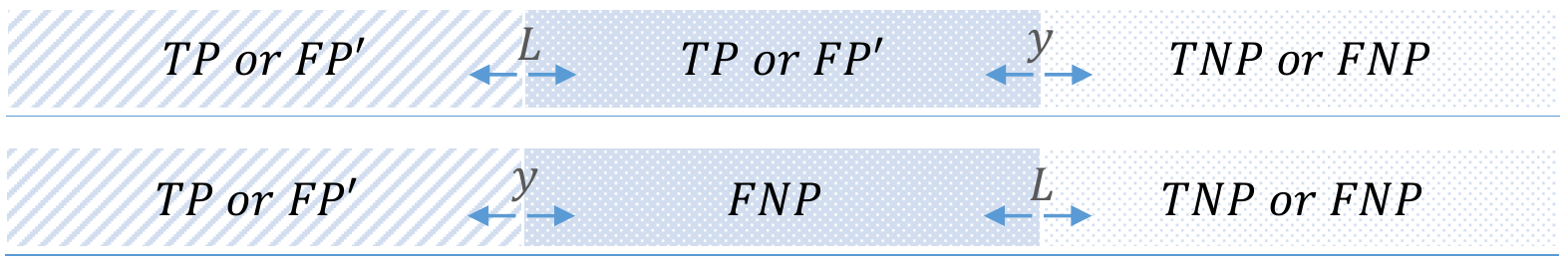}
\caption{Evaluation of our approach based on the chosen ranking (ordering of transformations)}
\label{evaluation:our}
\end{table}

\begin{compactenum}
\item{\textbf{if $y>L$} (i.e., we predict too many positive transformations), the current transformation in position $c$ can be below $y$, where we can find either $TP$ or $FP'$ and above $y$, where we can find either $TNP$ or $FNP$.}
\item{\textbf{if $y\leq L$} (i.e., we predict too few positive transformations), the current transformation in position $c$ can be: below $y$, between $y$ and $L$, and above $L$. Below $y$, we can find either $TP$ or $FP'$. Above $L$, we can find either $TNP$ or $FNP$. Finally, between $y$ and $L$ we can only find $FNP$, since here we have the transformations that are predicted as non-positive ($c>y$), but in reality they have positive impact,} 
\end{compactenum}
Using the aforementioned, we can calculate accuracy measures, where below the diagonal we can compute the ratio of true positives ($TP/(TP+FP'+TNP+FNP)$), and above the diagonal we can calculate the ratio of true non positives ($TNP/(TP+FP'+TNP+FNP)$). 

\begin{table*}[!b]
\centering
\includegraphics[width=0.9\textwidth]{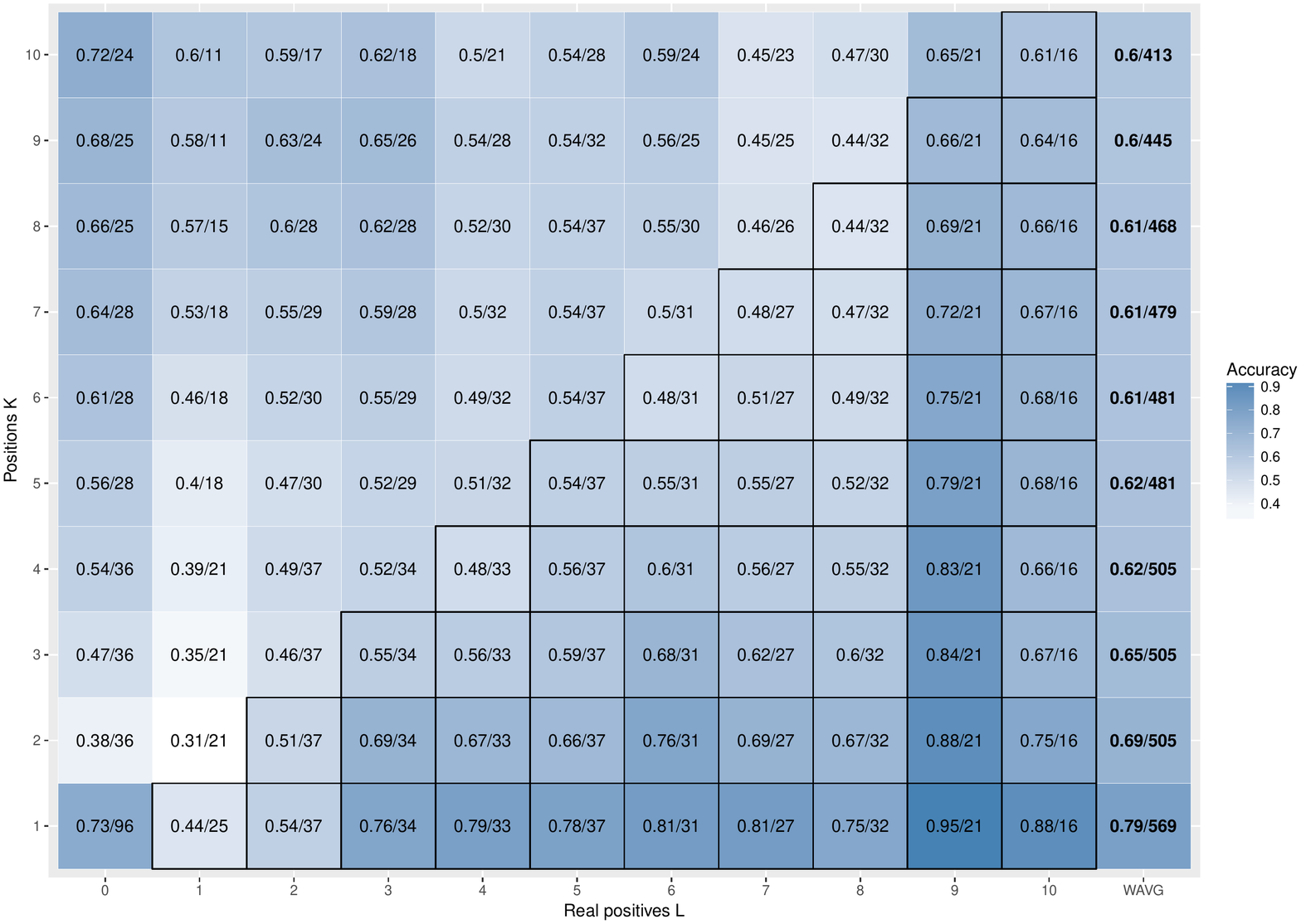}
\caption{Accuracy values obtained for the IBk classifier, for transformations in positions K in datasets with L real positive transformations. The numbers shown inside the cells, denoted as $x/z$, are the cumulative accuracies $x$, and the number of datasets $z$, which have at least $K$ transformations and exactly $L$ real positives. The last column, shows the values obtained after computing the weighted average for each position K for all possible L}
\label{evaluation:accuracy}
\end{table*}

In Table~\ref{evaluation:accuracy}, we show the results obtained for algorithm IBk. The numbers shown inside each cell $[L,K]$, denoted as $x/z$, are the cumulative accuracies $x$, and the number of datasets $z$, which have at least $K$ transformations and exactly $L$ real positives.
 
The cells are colored according to the values obtained, the darker the color the higher the accuracy. 

Notice that the cells below the diagonal become darker as $L$ grows. This means that we obviously perform better when the number of real positives is higher. Furthermore, the cells in the bottom part are darker than the rest. This means that we perform better in the first (top) $K$ positions. 

In Table~\ref{tbl:WAVG_K=1}, we show the weighted average results for all L, for the rest of the algorithms\footnote{The complete results for all the algorithms can be found in http://www.essi.upc.edu/dtim/people/bbilalli/KBS18 /results.}, but only for position K=1.
 
\begin{table}[!htbp]
\centering
\begin{tabular}{|l|c|}
\hline
Algorithm & WAVG in K=1     \\ \hline
J48             & 0.65/569     \\ \hline
Naive Bayes     & 0.80/569 \\ \hline
PART            & 0.69/569 \\ \hline
Logistic        & 0.59/512  \\ \hline
\end{tabular}
\caption{Weighted average values for all L in K=1}
\label{tbl:WAVG_K=1}
\end{table}

\subsubsection*{Comparison with a random pick}

To evaluate the performance of our approach, we compare it to the approach of a user randomly choosing a transformation to apply. For the latter, given the data, we need to find the probability of having TP below the diagonal and having TNP above the diagonal. 

Finding the probability of having TP below the diagonal in the $K^{th}$ position, translates to the problem of finding the probability of picking a positive transformation in $K$ draws, from a bag consisting of positive, negative, and zero (neutral) transformations. This follows a hyper-geometric distribution, and the expected value of TP in a cell $[L,K]$ below the diagonal is calculated as $\mu_{TP}=K\frac{L}{|T_d|}$,
where $|T_d|$ denotes the total number of transformations in dataset $d$. The expected value of TNP is calculated as $\mu_{TNP}=K\frac{L-|T_d|}{|T_d|}$. 

\begin{table}[!h]
\centering
\includegraphics[width=0.5\textwidth]{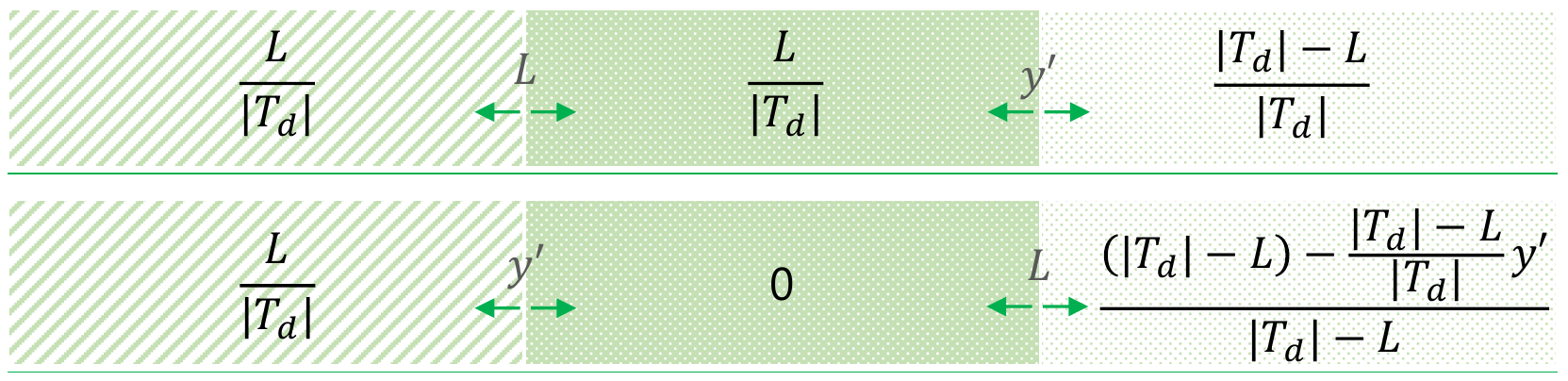}
\caption{Evaluation of the random pick based on the chosen ranking}
\label{evaluation:random}
\end{table}

To make a fair comparison with our approach, we need to assume the same ordering. Hence, the values to be calculated are shown in Table~\ref{evaluation:random}, where $y'$ is the expected number of positive predictions a dataset can have, which is calculated as the ratio of the transformations of dataset $d$, given the proportion of all real positives we can have for algorithm $a$, $y'=|T_d|P_a(Positive)$ (probabilities for each algorithm can be found using the distributions of the impacts in Figure~\ref{fig:impact_barPlot}). 

In Table~\ref{evaluation:random}, again two scenarios are considered:
\begin{compactenum}
\item{\textbf{if $y'>L$} (i.e., the random picks too many positive transformations), below $y'$ we take the probability of being $TP$, and above $y'$ we take the probability of being $TNP$.}
\item{\textbf{if $y'\leq L$} (i.e, the random picks too few positive transformations), for the current transformation $c$ below $y$, we calculate the probability of being $TP$. For the transformation positioned between $y'$ and $L$, the probability of being $TP$ is $0$. Finally, for the transformation above $L$, we calculate the expected $TNP$ based on the expected positive predictions $y'$,}
\end{compactenum}
More precisely, given the conditions in Table~\ref{evaluation:random}, the probabilities for each cell $[L,K]$, are calculated using the following function,

\begin{equation*}
  P(L,K)=\left\{
  \begin{array}{@{}ll@{}}
   (min(K,y')\frac{L}{|T_d|} + max(0,K-y')\frac{|T_d|-L}{|T_d|})/K, & \text{if}\ y'\geq{L}\\
    (min(K,y')\frac{L}{|T_d|} + max(0,K-L)\frac{(|T_d|-L)-(\frac{|T_d|-L}{|T_d|}y')}{|T_d|-L})/K, & \text{if}\ y'<{L}
  \end{array}\right.
\end{equation*}

To show whether the values obtained by our approach are significant, we performed a binomial distribution test comparing the true positives obtained by our approach with the total number of datasets with respect to the theoretical probabilities obtained by the random pick. The results obtained are shown in Table~\ref{evaluation:binom}, where the color of the cell denotes whether the value obtained is significant or not (white means significant). We consider the value to be significant if it is $p<=0.001$.

\begin{table*}[!h]
\centering
\includegraphics[width=0.9\textwidth]{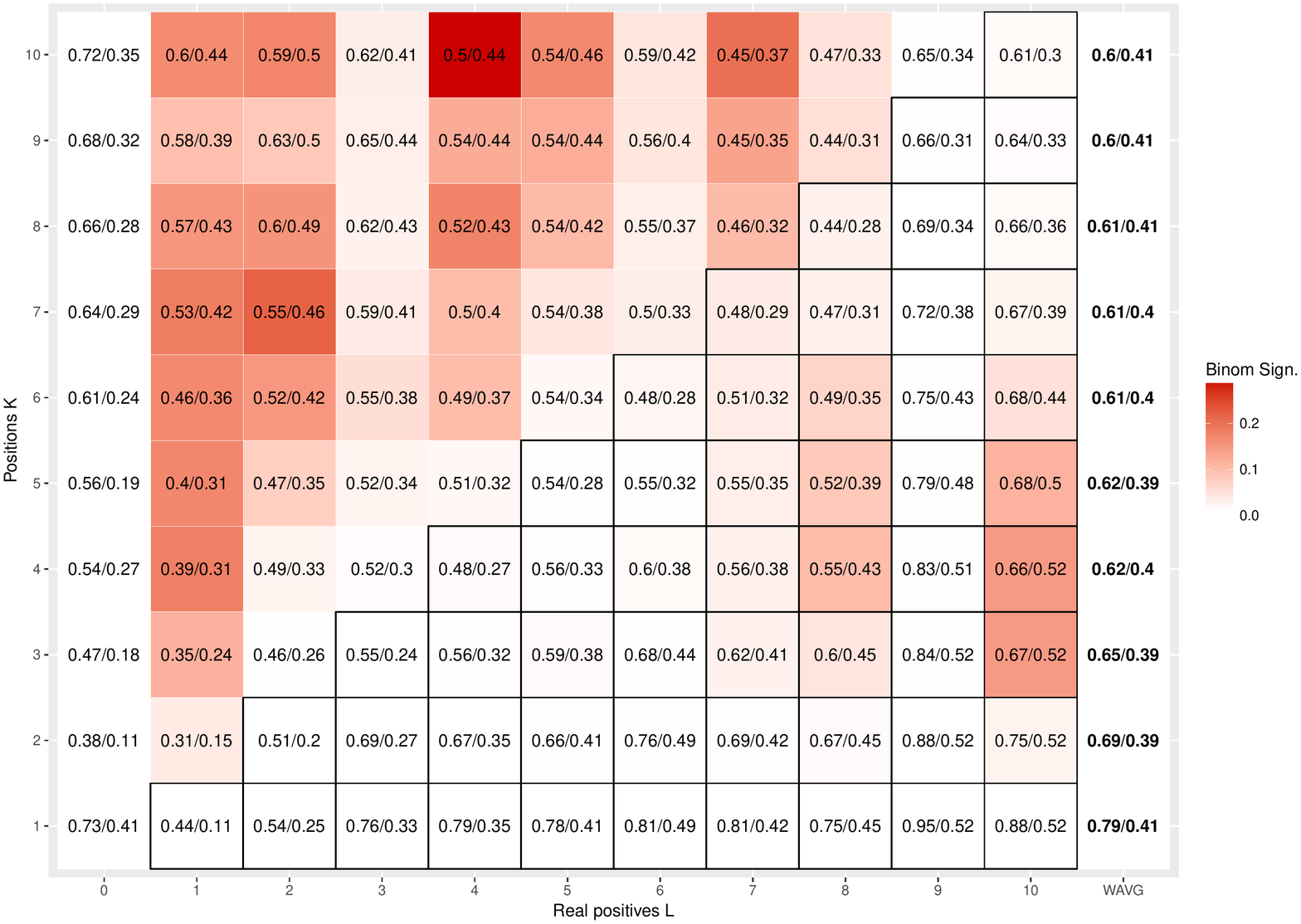}
\caption{Significance values obtained when comparing results obtained with meta-learning and the random pick, for the IBk classifier. The numbers shown inside the cells, denoted as $x/v$, are the cumulative accuracies $x$ obtained using our approach, and the random pick probabilities $z$, for datasets with at least $K$ transformations and exactly $L$ real positives. The last column shows the averages for our approach compared to the random pick, in each position $K$, weighted by the number of datasets in each L}
\label{evaluation:binom}
\end{table*}

It can be observed that significant values are obtained for most of the cells below the diagonal where the accuracy with regards to TP is measured. Furthermore, it is worth to mention that: 
\begin{compactitem}
    
\item[--]{observing the bottom left-most cell, it can be noted that the system almost doubles the chance of finding a transformation that does not positively impact the analysis (the system may suggest avoiding those transformations)}

\item[--]{the bottom right-most cell indicates that for position $K=1$, we almost double the accuracy compared to the random pick ($79\%$ versus $41\%$). Moreover, the accuracy obtained for the whole set of transformations for IBk was 67$\%$ and for top-1, becomes 79$\%$}

\item[--]{the probabilities of the random pick start to become higher above the diagonal, due to the fact that it is easier to guess negative transformations as you go down in the ranking}

\item[--]{significance values are also impacted by the sample sizes (number of datasets), which are different for each $L$ and they may also vary for different $K$s. Yet, observe the last column where the weighted averages are shown. The calculations are done for all the datasets on each $K$, and the values obtained are significant.}
\end{compactitem}

\subsection{Evaluation of the gain obtained from recommendations}

In the Information Retrieval domain, different measures that calculate the gain obtained from a ranked result have been proposed. The most popular one among them is the Discounted Cumulative Gain (DCG)~\cite{IRevalSIFIR00}.
The assumption is that the greater the ranked position, the less valuable the item (i.e., transformations) is for the user, because it is less likely that the user will examine it. Thus, DCG uses a discounting function that progressively reduces the gain as the rank increases. To compute DCG, a permutation/ordering $\pi$ of the gain values $G_{T_d}$ on the entire list of transformations in a given dataset $d$, results in the ordered list  of gains $G_{\pi,{T_d}}$ (a vector of length $N$, where $N$ is $|T_d|$). In particular, we are interested in the permutation that sorts according to our predicted scores (i.e., predicted probabilities) for the various transformations in a dataset: $G_{rec,{T_d}}$ denotes the recommended list or gains in descending order of the predicted scores of the transformations (highest ranked transformation is first in list). Moreover, the best ($G_{best,{T_d}}$) and worst ($G_{worst,{T_d}}$) possible permutations are also of interest.

For the general case DCG is computed as \cite{Steck13ACM}:
\begin{equation*}
    DCG_{G_{\pi,{T_d}}}=\sum_{i=1}^{N}\frac{G_{\pi,{T_d}}[i]}{log_2(i+1)}
\end{equation*}
 
The calculations are performed for each dataset and we obtain values for the recommended ranking ($DCG_{G_{rec,{T_d}}}$), the best ranking ($DCG_{G_{best,{T_d}}}$) and the worst ranking ($DCG_{G_{worst,{T_d}}}$). 
To obtain a relative value, we normalize the gain obtained by our recommendations in the following way: 
\begin{equation*}
    nDCG_d=\frac{DCG_{G_{rec,{T_d}}}-DCG_{G_{worst,{T_d}}}}{DCG_{G_{best,{T_d}}}-DCG_{G_{worst,{T_d}}}},
\end{equation*}
This normalized value can be interpreted as a percentage of how close we are to the best ranking, and it is calculated for each dataset. Averaging the individual measures over all datasets with at least one relevant (non-neutral) transformation we obtain the mean $\overline{nDCG}$. This measure can again be calculated for the whole set of transformations or for the top-K. In Table~\ref{tbl:DCG}, we provide the results obtained for the whole set of transformations and for top-1, for all the classification algorithms considered. 

\begin{table}[!htbp]
\begin{minipage}[c]{\textwidth}
\centering
\begin{tabular}{|l|c|c|c|}
\hline
\multicolumn{1}{|l|}{\multirow{2}{*}{Algorithm}} & \multicolumn{2}{c|}{$\overline{nDCG}$} & \multicolumn{1}{c|}{\multirow{2}{*}{\#Datasets considered\footnote{Number of datasets with at least 1 relevant (non-neutral) transformation}}}\\ \cline{2-3} 
\multicolumn{1}{|c|}{}                           & All trans.     & Top-1 &      \\ \hline
J48             & 0.73     &    0.79    &  503  \\ \hline
Naive Bayes     & 0.78     &    0.84    &  504  \\ \hline
PART            & 0.73     &    0.79    &  503  \\ \hline
Logistic        & 0.64     &    0.67    &  447  \\ \hline
IBk             & 0.77     &    0.84    &  500  \\ \hline
\end{tabular}
\end{minipage}
\caption{Normalized discounted cumulative gain values}
\label{tbl:DCG}
\end{table}

\section{Related work}

A lot of research has been done in order to address the problem of providing user support in the different steps of KDD. 

Initially, the focus has been on developing (semi) automatic systems, to provide user assistance in one or many steps altogether. These systems have been referred to as Intelligent Discovery Assistants (IDAs). Recently however, the direction has shifted towards designing systems that specifically provide user assistance in the data pre-processing step. 
In the following, we give more details about both types of the aforementioned systems.

\noindent\textbf{Intelligent Discovery Assistants}. IDAs are grouped into the following main categories~\cite{Serban13survey}: Expert systems, Meta-learning systems, Case-based reasoning systems, Planning-based data analysis systems. 

\emph{Expert systems} \cite{Raes92springex,Sleeman95consultant}. They are the first and simplest systems to provide help to the user during the data mining phase. Their main component is a knowledge base consisting of expert rules. Given the input from the user, the rules are used to determine the mining algorithms to be recommended. 

\emph{Meta-learning systems (MLS)} \cite{KalousisNoemon01,Giraud05dma}. They are more advanced than the previous ones. The rules that were statically defined by the experts in the previous category are dynamically learned here. MLSs try to discover the relationship between measurable features of the dataset and the performance of different algorithms, which is a standard learning problem. The learned model is then used to predict the most suitable data mining algorithm for a given dataset. 

\emph{Case-based reasoning systems (CBS)} \cite{Engels96planning,Lindner99ast,Morik02miningmart}. They store the successfully applied workflows as cases, in a \emph{case base}, with the goal of reusing them in the future. When faced with a new problem (i.e., dataset) provided by the user, these systems return $k$ similar cases, which can be further adapted to the current problem.

\emph{Planning-based data analysis systems (PDA)} \cite{Diamantini09kddvm,Zakova11rdm,Kietz14eIDA}. PDAs are able to autonomously design valid workflows without relying on similarities. To this end, the workflow composition problem is treated as a planning problem, 
where the input, output, preconditions, and effects (IOPE) of each operator (algorithm) need to be formally defined. Multiple workflows are then generated by combining operators that complement one another.

The drawback of IDAs is that they mainly focus on providing user support in the data mining step. An exception is MiningMart from the CBR category, where the cases are workflows containing also data pre-processing operators. However, the problem is that the user needs to find a similar case which may not exist and further adapt it for its needs.
Furthermore, PDAs also consider data pre-processing yet they do not address the problem of ranking the explosive number of generated workflows. Hence, although they may recommend semantically valid workflows, those workflows may be far from optimal.

\noindent\textbf{User assistance for data pre-processing.}
Since pre-processing covers a broad range of activities, different systems have been developed to tackle the problem of user assistance from different perspectives.
There are systems that discover patterns and detect errors in data and then automatically infer relevant transformations. For instance, in Potter's Wheel \cite{PottersWheel01VLDB}, Wrangler \cite{WranglerCHI11}, and Foofah \cite{Foofah17SIGMOD} the relevant transformations are learned by example. The user either directly manipulates the visualized data or he/she needs to provide the output (to be) data. 

In KATARA \cite{Katara15SIGMOD}, DataXFormer \cite{DataXFormer15SIGMOD}, and VADA \cite{VADA17SIGMOD}, they also infer transformations, however, this time using external knowledge stored in knowledge bases, web tables, or even by knowledge obtained interacting with crowds.

Other systems like NADEEF \cite{NADEEF13SIGMOD}, Llunatic \cite{Llunatic13VLDBEnd}, and BigDansing \cite{BigDansing15SIGMOD}, (semi) automate the detection and repairing of violations with respect to a set of heterogeneous and ad-hoc constraints. Many types of quality constraints like functional dependencies, conditional functional dependencies, multivalue dependencies, and ETL rules can be defined. Their goal is to cope with multiple queries holistically and optimize their application. 

In DataTamer \cite{DataTamer13CIDR} and DataCivilizer \cite{DataCivilizer17CIDR}, they deal with the end to end curation (e.g., integration, de-duplication) of data from different sources.

Finally, in ActiveClean \cite{ActiveClean16SIGMOD}, they aim at prioritizing the cleaning of records that are more likely to affect the results of the statistical modeling problems, assuming that the latter belong to the class of models called convex loss models (e.g., linear regression and SVMs). 

Note that the assumption of the aforementioned systems is that an expert user is performing the analysis. That is, the user knows what the final shape of the transformed data should look like. Our assumption is that the user is a non-expert, and he/she wants to apply transformations only for the sake of improving the analysis. That is, he/she does not know what the input dataset to a mining algorithm should look like, in order to yield better results/analysis.

\section{Conclusion and future work}


In this work, we addressed the problem of assisting non-expert users to perform pre-processing with the goal of improving the final results of their classification tasks.

To provide assistance, we trained a model that learned the relationship between pre-processing operators and the performance of classification algorithms. To this end, we were able to rank the transformations according to their impact on the final result of the analysis (i.e., the impact of transformations on the predictive accuracy of a classification algorithm). An extensive evaluation on hundreds of datasets and a set of classification algorithms, showed that our approach gives promising results. More specifically, we were able to observe that:
\begin{compactitem} 
    \item[--] even if a user randomly picks a transformation from the entire list of transformations we obtain an average accuracy of 63\%, for all the algorithms considered
    \item[--] recommending only the top-1 transformation, increased the accuracy to 71\% on average 
    \item[--] measuring the gain obtained from our ranking for all transformations using DCG, we found that we are as close as 73\%, on average to the gain obtained from the best possible ranking (for all the algorithms considered)
    \item[--] measuring the gain from the top-1 recommendations using DCG, we found that we were as close as 79\%, on average to the gain obtained from the best possible ranking
\end{compactitem}
Finally, the results indicate that our tool PRESISTANT, can assist users to more effectively identify the pre-processing operators appropriate to their applications, and to achieve improved results.

As a future work, we see potential value on extending the list of the pre-processing operators and classification algorithms we have considered so far and we also plan to incorporate regression algorithms.

\bibliographystyle{elsarticle-num}

\noindent\textbf{Acknowledgments.} This research has been funded by the European Commission through the Erasmus Mundus Joint Doctorate ``Information Technologies for Business Intelligence - Doctoral College" (IT4BI-DC).

\section*{References}
\bibliography{references}

\end{document}